\documentclass[11pt,a4paper]{article}
\usepackage[blocks]{authblk}
\usepackage[hyperref]{acl2018}
\usepackage{times}
\usepackage{latexsym}
\usepackage{amsmath}
\usepackage{graphicx}
\usepackage{url}

\aclfinalcopy % Uncomment this line for the final submission
\def\aclpaperid{95} %  Enter the acl Paper ID here

\title{Neural Open Information Extraction}

\author{\textbf{Lei Cui}}
\author{\textbf{Furu Wei}}
\author{\textbf{Ming Zhou}}
\affil{Microsoft Research Asia \authorcr \tt\{lecu,fuwei,mingzhou\}@microsoft.com}

\date{}

\begin{document}
\maketitle
\begin{abstract}
Conventional Open Information Extraction (Open IE) systems are usually built on hand-crafted patterns from other NLP tools such as syntactic parsing, yet they face problems of error propagation. In this paper, we propose a neural Open IE approach with an encoder-decoder framework. Distinct from existing methods, the neural Open IE approach learns highly confident arguments and relation tuples bootstrapped from a state-of-the-art Open IE system. An empirical study on a large benchmark dataset shows that the neural Open IE system significantly outperforms several baselines, while maintaining comparable computational efficiency.

\end{abstract}

\section{Introduction}

Open Information Extraction (Open IE) involves generating a structured representation of information in text, usually in the form of triples or n-ary propositions. An Open IE system not only extracts arguments but also relation phrases from the given text, which does not rely on pre-defined ontology schema. For instance, given the sentence ``\textit{deep learning is a subfield of machine learning}'', the triple (\textit{deep learning}; \textit{is a subfield of}; \textit{machine learning}) can be extracted, where the relation phrase ``\textit{is a subfield of}'' indicates the semantic relationship between two arguments. Open IE plays a key role in natural language understanding and fosters many downstream NLP applications such as knowledge base construction, question answering, text comprehension, and others. 

The Open IE system was first introduced by \textsc{Text}\textsc{Runner}~\cite{Banko:2007:OIE:1625275.1625705}, followed by several popular systems such as \textsc{Re}\textsc{Verb}~\cite{ReVerb2011}, \textsc{Ollie}~\cite{ollie-emnlp12}, ClausIE~\cite{DelCorro:2013:CCO:2488388.2488420} Stanford \textsc{Open}IE~\cite{angeli-johnsonpremkumar-manning:2015:ACL-IJCNLP}, PropS~\cite{DBLP:journals/corr/StanovskyFDG16} and most recently \textsc{Open}IE4\footnote{https://github.com/allenai/openie-standalone}~\cite{Mausam:2016:OIE:3061053.3061220} and \textsc{Open}IE5\footnote{https://github.com/dair-iitd/OpenIE-standalone}. Although these systems have been widely used in a variety of applications, most of them were built on hand-crafted patterns from syntactic parsing, which causes errors in propagation and compounding at each stage~\cite{Banko:2007:OIE:1625275.1625705, gashteovski-gemulla-delcorro:2017:EMNLP2017,schneider-EtAl:2017:BLGNLP2017}. Therefore, it is essential to solve the problems of cascading errors to alleviate extracting incorrect tuples.

\begin{figure*}[ht]
\center
  \includegraphics[width=0.95\textwidth]{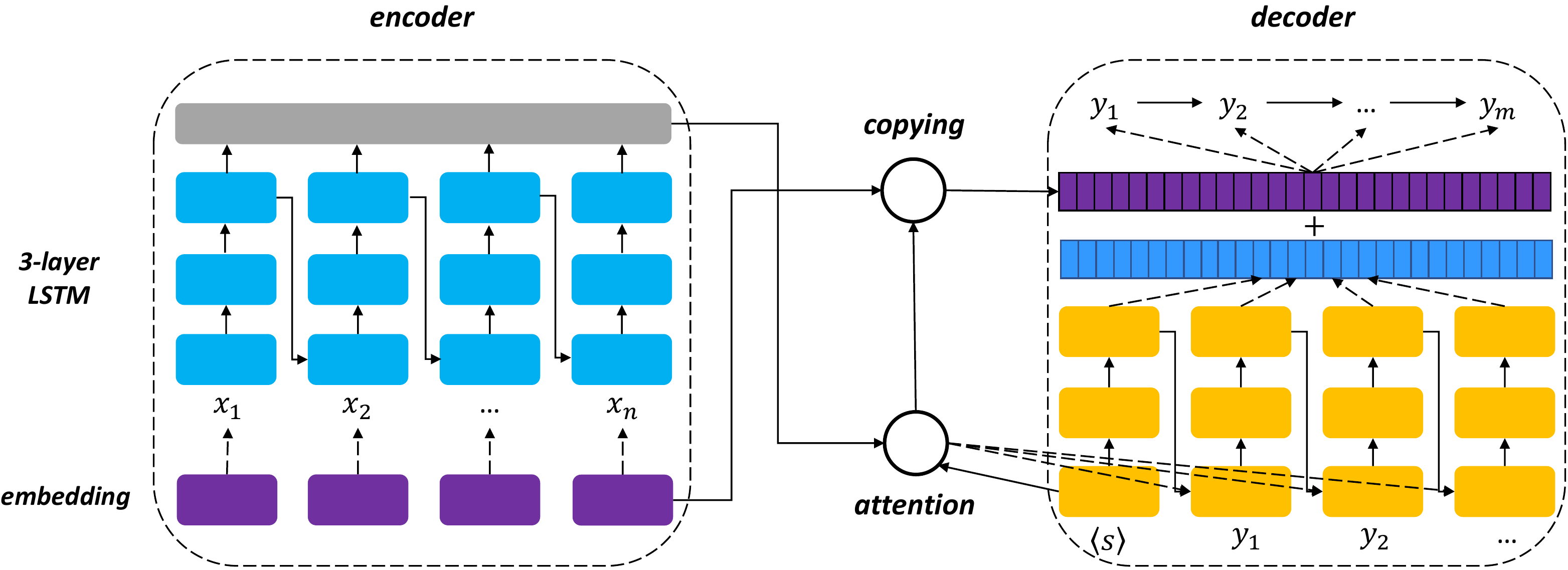}\\
  \caption{The encoder-decoder model architecture for the neural Open IE system}\label{1:fig}
\end{figure*}

To this end, we propose a neural Open IE approach with an encoder-decoder framework. The encoder-decoder framework is a text generation technique and has been successfully applied to many tasks, such as machine translation~\cite{cho-EtAl:2014:EMNLP2014, DBLP:journals/corr/BahdanauCB14, DBLP:journals/corr/SutskeverVL14, DBLP:journals/corr/WuSCLNMKCGMKSJL16, gehring2017convs2s, NIPS2017_7181}, image caption~\cite{DBLP:journals/corr/VinyalsTBE14}, abstractive summarization~\cite{rush-chopra-weston:2015:EMNLP, Nallapati2016AbstractiveTS, see-liu-manning:2017:Long} and recently keyphrase extraction~\cite{meng-EtAl:2017:Long}. Generally, the encoder encodes the input sequence to an internal representation called `context vector' which is used by the decoder to generate the output sequence. The lengths of input and output sequences can be different, as there is no one on one relation between the input and output sequences. In this work, Open IE is cast as a sequence-to-sequence generation problem, where the input sequence is the sentence and the output sequence is the tuples with special placeholders. For instance, given the input sequence ``\textit{deep learning is a subfield of machine learning}'', the output sequence will be ``$\langle$arg1$\rangle$ deep learning $\langle$/arg1$\rangle$ $\langle$rel$\rangle$ is a subfield of  $\langle$/rel$\rangle$ $\langle$arg2$\rangle$ machine learning $\langle$/arg2$\rangle$''. We obtain the input and output sequence pairs from highly confident tuples bootstrapped from a state-of-the-art Open IE system. Experiment results on a large benchmark dataset illustrate that the neural Open IE approach is significantly better than others in precision and recall, while also reducing the dependencies on other NLP tools.

The contributions of this paper are threefold. First, the encoder-decoder framework learns the sequence-to-sequence task directly, bypassing other hand-crafted patterns and alleviating error propagation. Second, a large number of high-quality training examples can be bootstrapped from state-of-the-art Open IE systems, which is released for future research. Third, we conduct comprehensive experiments on a large benchmark dataset to compare different Open IE systems to show the neural approach's promising potential. 

%The rest of the paper is organized as follows: The proposed approach is explained in Section 2. Experimental results are presented in Section 3. Section 4 introduces some related work. Section 5 concludes the paper and suggests future research directions.

\section{Methodology}

\subsection{Problem Definition}

Let $(X, Y)$ be a sentence and tuples pair, where $X=(x_1,x_2,...,x_m)$ is the word sequence and $Y=(y_1,y_2,...,y_n)$ is the tuple sequence extracted from $X$. The conditional probability of $P(Y|X)$ can be decomposed as:
\begin{equation}
\begin{split}
P(Y|X) & = P(Y|x_1,x_2,...,x_m) \\
& = \prod_{i=1}^{n}p(y_i|y_1,y_2,...,y_{i-1};x_1,x_2,...x_m)
\end{split}
\end{equation}
In this work, we only consider the binary extractions from sentences, leaving n-ary extractions and nested extractions for future research. In addition, we ensure that both the argument and relation phrases are sub-spans of the input sequence. Therefore, the output vocabulary equals the input vocabulary plus the placeholder symbols.

\subsection{Encoder-Decoder Model Architecture}

The encoder-decoder framework takes a variable length input sequence to a compressed representation vector that is used by the decoder to generate the output sequence. In this work, both the encoder and decoder are implemented using Recurrent Neural Networks (RNN) and the model architecture is shown in Figure \ref{1:fig}.

The encoder uses a 3-layer stacked Long Short-Term Memory (LSTM)~\cite{Hochreiter:1997:LSM:1246443.1246450} network to covert the input sequence $X=(x_1,x_2,...x_m)$ into a set of hidden representations $\textbf{h}=(h_1,h_2,...,h_m)$, where each hidden state is obtained iteratively as follows:
\begin{equation}
h_t = \textbf{LSTM}(x_t, h_{t-1})
\end{equation}

The decoder also uses a 3-layer LSTM network to accept the encoder's output and generate a variable-length
sequence $Y$ as follows:
\begin{equation}
\begin{split}
s_t &= \textbf{LSTM}(y_{t-1},s_{t-1},c)\\
p(y_t)&=\textbf{softmax}(y_{t-1},s_t,c)
\end{split}
\end{equation}
where $s_t$ is the hidden state of the decoder LSTM at time $t$, $c$ is the context vector that is introduced later. We use the softmax layer to calculate the output probability of $y_t$ and select the word with the largest probability.

An attention mechanism is vital for the encoder-decoder framework, especially for our neural Open IE system. Both the arguments and relations are sub-spans that correspond to the input sequence. We leverage the attention method proposed by~\citeauthor{DBLP:journals/corr/BahdanauCB14} to calculate the context vector $c$ as follows:
\begin{equation}
\begin{split}
c_i &= \sum_{j=1}^{n}\alpha_{ij}h_j\\
\alpha_{ij} &= \frac{\text{exp}(e_{ij})}{\sum_{k=1}^{n}\text{exp}(e_{ik})} \\
e_{ij} &= a(s_{i-1}, h_j)
\end{split}
\end{equation}
where $a$ is an alignment model that scores how well the inputs around position $j$ and the output at position $i$ match, which is measured by the encoder hidden state $h_j$ and the decoder hidden state $s_{i-1}$. The encoder and decoder are jointly optimized to maximize the log probability of the output sequence conditioned on the input sequence.

\subsection{Copying Mechanism}

Since most encoder-decoder methods maintain a fixed vocabulary of frequent words and convert a large number of long-tail words into a special symbol ``$\langle$unk$\rangle$'', the copying mechanism~\cite{gu-EtAl:2016:P16-1,gulcehre-EtAl:2016:P16-1,see-liu-manning:2017:Long,meng-EtAl:2017:Long} is designed to copy words from the input sequence to the output sequence, thus enlarging the vocabulary and reducing the proportion of generated unknown words. For the neural Open IE task, the copying mechanism is more important because the output vocabulary is directly from the input vocabulary except for the placeholder symbols. We simplify the copying method in~\cite{see-liu-manning:2017:Long}, the probability of generating the word $y_t$ comes from two parts as follows:
\begin{equation}
p(y_t)=
\begin{cases}
    p(y_t|y_1,y_2,...,y_{t-1};X) & \text{if $y_t\in V$} \\
    \sum_{i:x_i=y_t}a_i^t & \text{otherwise} \\
  \end{cases}
\end{equation}
where $V$ is the target vocabulary. We combine the sequence-to-sequence generation and attention-based copying together to derive the final output.

\section{Experiments}

\subsection{Data}

For the training data, we used Wikipedia dump 20180101\footnote{https://dumps.wikimedia.org/enwiki/20180101/} and extracted all the sentences that are 40 words or less. \textsc{Open}IE4 is used to analyze the sentences and extract all the tuples with binary relations. To further obtain high-quality tuples, we only kept the tuples whose confidence score is at least 0.9. Finally, there are a total of 36,247,584  $\langle$sentence, tuple$\rangle$ pairs extracted. The training data is released for public use at \url{https://1drv.ms/u/s!ApPZx_TWwibImHl49ZBwxOU0ktHv}.

For the test data, we used a large benchmark dataset~\cite{stanovsky-dagan:2016:EMNLP2016} that contains 3,200 sentences with 10,359 extractions\footnote{\url{https://github.com/gabrielStanovsky/oie-benchmark}}. We compared with several state-of-the-art baselines including \textsc{Ollie}, ClausIE, Stanford \textsc{Open}IE, PropS and \textsc{Open}IE4. The evaluation metrics are precision and recall.

\subsection{Parameter Settings}

We implemented the neural Open IE model using OpenNMT~\cite{DBLP:journals/corr/KleinKDSR17}, which is an open source encoder-decoder framework. We used 4 M60 GPUs for parallel training, which takes 3 days. The encoder is a 3-layer bidirectional LSTM and the decoder is another 3-layer LSTM. Our model has 256-dimensional hidden states and 256-dimensional word embeddings. A vocabulary of 50k words is used for both the source and target sides. We optimized the model with SGD and the initial learning rate is set to 1. We trained the model for 40 epochs and started learning rate decay from the $\text{11}^{th}$ epoch with a decay rate 0.7. The dropout rate is set to 0.3. We split the data into 20 partitions and used data sampling in OpenNMT to train the model. This reduces the length of the epochs for more frequent learning rate updates and validation perplexity computation.

\subsection{Results}

\begin{figure*}[ht]
\center
  \includegraphics[width=0.99\textwidth]{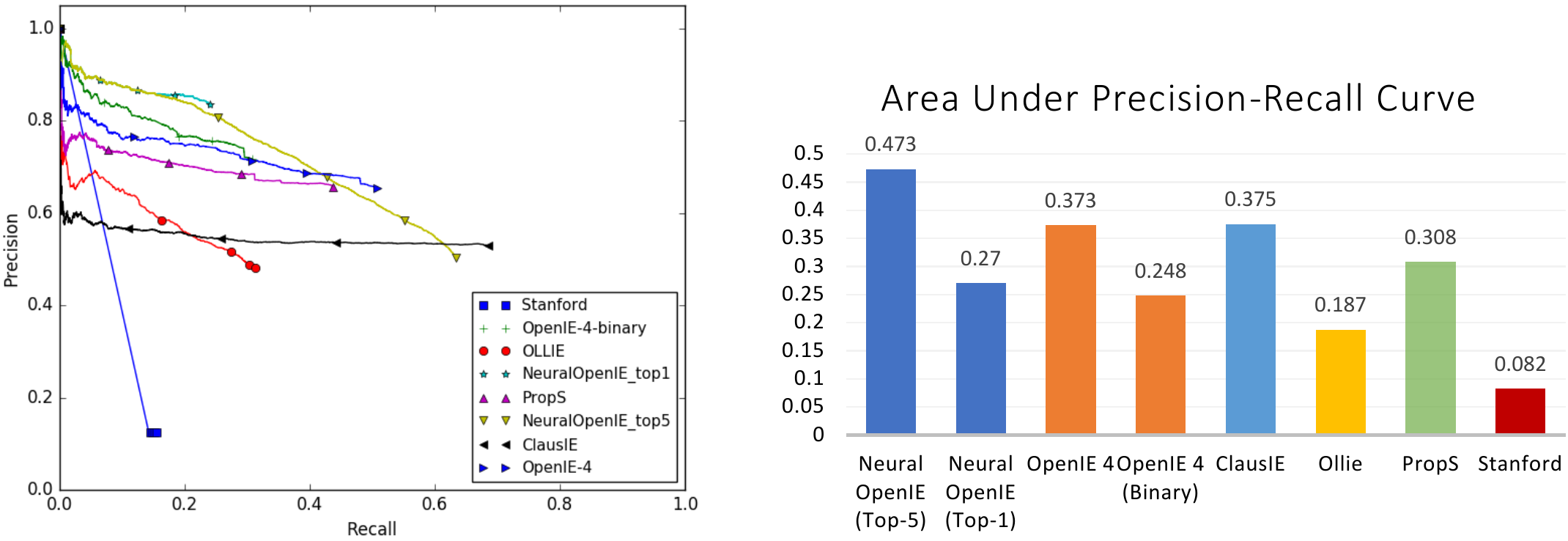}\\
  \caption{The Precision-Recall (P-R) curve and Area under P-R curve (AUC) of Open IE systems}\label{2:fig}
\end{figure*}

We used the script in~\cite{stanovsky-dagan:2016:EMNLP2016}\footnote{The absolute scores are different from the original paper because the authors changed the matching function in their GitHub Repo, but did not change the relative performance.} to evaluate the precision and recall of different baseline systems as well as the neural Open IE system. The precision-recall curve is shown in Figure~\ref{2:fig}. It is observed that the neural Open IE system performs best among all tested systems. Furthermore, we also calculated the Area under Precision-Recall Curve (AUC) for each system. The neural Open IE system with top-5 outputs achieves the best AUC score 0.473, which is significantly better than other systems. Although the neural Open IE is learned from the bootstrapped outputs of \textsc{Open}IE4's extractions, only 11.4\% of the extractions from neural Open IE agree with the \textsc{Open}IE4's extractions, while the AUC score is even better than \textsc{Open}IE4's result. We believe this is because the neural approach learns arguments and relations across a large number of highly confident training instances. This also indicates that the generalization capability of the neural approach is better than previous methods. We observed many cases in which the neural Open IE is able to correctly identify the boundary of arguments but \text{Open}IE4 cannot, for instance:

\begin{table}[h]
\small
\begin{center}
\begin{tabular}{l p{5cm}}
\hline
\bf{Input} &  Instead , much of numerical analysis is concerned with obtaining approximate solutions while maintaining reasonable bounds on errors . \\\hline\hline
\bf{Gold} & much of numerical analysis $|||$ concerned $|||$ with obtaining approximate solutions \bf{while maintaining reasonable bounds on errors}\\\hline
\bf{OpenIE4} & much of numerical analysis $|||$ is concerned with $|||$ obtaining approximate solutions\\\hline
\bf{Neural Open IE} & much of numerical analysis $|||$ is concerned $|||$ with obtaining approximate solutions \bf{while maintaining reasonable bounds on errors}\\\hline
\end{tabular}
\end{center}
\label{tab:example}
\end{table}
\noindent This case illustrates that the neural approach reduces the limitation of hand-crafted patterns from other NLP tools. Therefore, it reduces the error propagation effect and performs better than other systems especially for long sentences. 

We also investigated the computational cost of different systems. For the baseline systems, we obtained the Open IE extractions using a Xeon 2.4 GHz CPU. For the neural Open IE, we evaluated performance based on an M60 GPU. The running time was calculated by extracting Open IE tuples from the test dataset that contains a total of 3,200 sentences. The results are shown in Table~\ref{1:tab}. Among the aforementioned conventional systems, Ollie is the most efficient approach which takes around 160s to finish the extraction. By using GPU, the neural approach takes 172s to extract the tuples from the test data, which is comparable with conventional approaches. As the neural approach does not depend on other NLP tools, we can further optimize the computational cost in future research efforts.

\begin{table}[h]
\begin{center}
\begin{tabular}{|c|c|c|}
\hline \textbf{System} & \textbf{Device} & \textbf{Time} \\\hline
Stanford & CPU & 234s\\\hline
Ollie & CPU & 160s \\\hline
ClausIE & CPU & 960s \\\hline
PropS & CPU & 432s \\\hline
OpenIE4 & CPU & 181s \\\hline
%Neural Open IE & CPU & \\\hline
Neural Open IE & GPU & 172s \\\hline
%Neural Open IE & 4 $\times$ GPU & \\\hline
\end{tabular}
\end{center}
\caption{Running time of different systems}
\label{1:tab}
\end{table}

\section{Related Work}

The development of Open IE systems has witnessed rapid growth during the past decade~\cite{Mausam:2016:OIE:3061053.3061220}. The Open IE system was introduced by \textsc{Text}\textsc{Runner}~\cite{Banko:2007:OIE:1625275.1625705} as the first generation. It casts the argument and relation extraction task as a sequential labeling problem. The system is highly scalable and extracts facts from large scale web content. \textsc{Re}\textsc{Verb}~\cite{ReVerb2011} improved over \textsc{Text}\textsc{Runner} with syntactic and lexical constraints on binary relations expressed by verbs, which more than doubles the area under the precision-recall curve. Following these efforts, the second generation known as R2A2~\cite{Etzioni:2011:OIE:2283396.2283398} was developed based on \textsc{Re}\textsc{Verb} and an argument identifier, \textsc{Arg}\textsc{Learner}, to better extract the arguments for the relation phrases. The first and second generation Open IE systems extract only relations that are mediated by verbs and ignore contexts. To alleviate these limitations, the third generation \textsc{Ollie}~\cite{ollie-emnlp12} was developed, which achieves better performance by extracting relations mediated by nouns, adjectives, and more. In addition, contextual information is also leveraged to improve the precision of extractions. All the three generations only consider binary extractions from the text, while binary extractions are not always enough for their semantics representations. Therefore, \textsc{Srl}\textsc{Ie}~\cite{christensen-EtAl:2010:FAMLBR} was developed to include an attribute context with a tuple when it is available. \textsc{Open}IE4 was built on \textsc{Srl}\textsc{Ie} with a rule-based extraction system \textsc{Rel}\textsc{Noun}~\cite{pal-:2016:W16-13} for extracting noun-mediated relations. Recently, \textsc{Open}IE5 improved upon extractions from numerical sentences~\cite{saha-pal-mausam:2017:Short} and broke conjunctions in arguments to generate multiple extractions. During this period, there were also some other Open IE systems emerged and successfully applied in different scenarios, such as ClausIE~\cite{DelCorro:2013:CCO:2488388.2488420} Stanford \textsc{Open}IE~\cite{angeli-johnsonpremkumar-manning:2015:ACL-IJCNLP}, PropS~\cite{DBLP:journals/corr/StanovskyFDG16}, and more.

The encoder-decoder framework was introduced by~\citeauthor{cho-EtAl:2014:EMNLP2014} and~\citeauthor{DBLP:journals/corr/SutskeverVL14}, where a multi-layered LSTM/GRU is used to map the input sequence to a vector of a fixed dimensionality, and then another deep LSTM/GRU to decode the target sequence from the vector.~\citeauthor{DBLP:journals/corr/BahdanauCB14} and ~\citeauthor{luong-pham-manning:2015:EMNLP} further improved the encoder-decoder framework by integrating an attention mechanism so that the model can automatically find parts of a source sentence that are relevant to predicting a target word. To improve the parallelization of model training, convolutional sequence-to-sequence (ConvS2S) framework~\cite{gehring2016convenc,gehring2017convs2s} was proposed to fully parallelize the training since the number of non-linearities is fixed and independent of the input length. Recently, the transformer framework~\cite{NIPS2017_7181} further improved over the vanilla S2S model and ConvS2S in both accuracy and training time.

In this paper, we use the LSTM-based S2S approach to obtain binary extractions for the Open IE task. To the best of our knowledge, this is the first time that the Open IE task is addressed using an end-to-end neural approach, bypassing the hand-crafted patterns and alleviating error propagation.

\section{Conclusion and Future Work}

We proposed a neural Open IE approach using an encoder-decoder framework. The neural Open IE model is trained with highly confident binary extractions bootstrapped from a state-of-the-art Open IE system, therefore it can generate high-quality tuples without any hand-crafted patterns from other NLP tools. Experiments show that our approach achieves very promising results on a large benchmark dataset.

For future research, we will further investigate how to generate more complex tuples such as n-ary extractions and nested extractions with the neural approach. Moreover, other frameworks such as convolutional sequence-to-sequence and transformer models could apply to achieve better performance.

\section*{Acknowledgments}

We are grateful to the anonymous reviewers for their insightful comments and suggestions.

% The acknowledgments should go immediately before the references.  Do not number the acknowledgments section ({\em i.e.}, use \verb|\section*| instead of \verb|\section|). Do not include this section when submitting your paper for review.

%\bibliography{acl2018}
\bibliography{acl2018}
\bibliographystyle{acl_natbib}

\end{document}